\documentclass[letterpaper, 10pt, conference]{ieeeconf}
\IEEEoverridecommandlockouts
\usepackage{graphics}
\usepackage{epsfig}
\usepackage{mathptmx}
\usepackage{times}
\usepackage{amsmath}
\usepackage{amssymb}
\usepackage{graphicx}
\usepackage{subcaption}
\usepackage{url}
\usepackage{tikz}
\usepackage{censor}
\usepackage{array}

\title{\LARGE \bf Enabling Tidying Tasks in Robotic Vacuum Cleaners\\ via a Geometry-Induced Self-Locking Deployable Linkage}

\author{Eunseok Song, Jinwoong Kim, Sohee Lim, Taehyeong Kim, Yongkook Kim, \\
Hyeontaek Lim, Hyungmin Son and Seungyong Hyung$^{*}$
\thanks{This work was conducted in October 2025 at Samsung Research, Samsung Electronics, 56, Seongchon-gil, Seocho-gu, Seoul, South Korea.}%
\thanks{E-mail: {\tt\footnotesize \{euns.song, seungyong.hyung\}@samsung.com}}
\thanks{* Corresponding author.}
}

\begin{document}

\maketitle
\thispagestyle{empty}
\pagestyle{empty}

\begin{abstract}
Robotic vacuum cleaners must extend their functionality within tight limits on size, cost, and durability. Deployable linkages with few actuators meet these limits, but their backdrivability makes it difficult to hold a load-bearing posture without continuous motor torque. We introduce a two-stage deployable linkage that locks through its own geometry. At full deployment, an internal three-link subchain approaches a collinear, near-singular configuration in which the input--output transmission ratio tends to zero; plate loads therefore leave only a small residual back-driving torque, which drive-train friction and a lightweight magnetic detent absorb. A single actuator performs deployment and locking, a passive chassis-mounted trigger releases the lock for retraction, and a second actuator adjusts only the deployment height. We integrate the mechanism into a commercial robotic vacuum cleaner and validate it through a 100-cycle repeatability test, a quasi-static load test in which the locked plate held 50\,N with the motor de-energized, and 140 pushing-type tidying trials on seven household objects against a commercial arm-equipped vacuum robot. We derive the analytical holding condition and discuss its limits as an intermediate design point between passive compliance and articulated manipulation.
\end{abstract}

\section{INTRODUCTION}\label{sec:intro}
Robotic vacuum cleaners are among the most widely deployed domestic robots \cite{forlizzi2006service}, and recent studies emphasize that they could be repurposed for simple assistance tasks during their substantial idle time \cite{shiokawa2025beyond, shafiullah2023bringing}. Any such extension, however, must fit within a commercial product's fixed footprint, low-profile chassis, battery budget, and bill-of-materials cost. Adding a multi-DoF arm is possible, as recent products and research systems for household tidying \cite{wu2023tidybot} and compact mobile manipulation \cite{kemp2022stretch} demonstrate, but it brings grasp planning, payload limits, and cost that are disproportionate to the simple pushing and containment actions that most floor-level tidying requires.

A low-cost alternative is a deployable linkage with few actuators. Such designs reduce actuation cost and, in the underactuated case, provide passive adaptability \cite{he2019underactuated, in2015exo, kontoudis2015open, badeau2018intuitive, chang2020soft, li2023linkage}. Their well-known weakness is the converse: a backdrivable linkage cannot hold a posture under load without continuous actuator torque, which wastes battery energy, and it may collapse or deviate when pushing against objects \cite{plooij2015lock, chen2020analysis}. The design challenge is therefore to obtain a rigid, load-bearing configuration on demand without adding locking hardware.

\begin{figure}[t]
\centering
\includegraphics[width=0.99\linewidth]{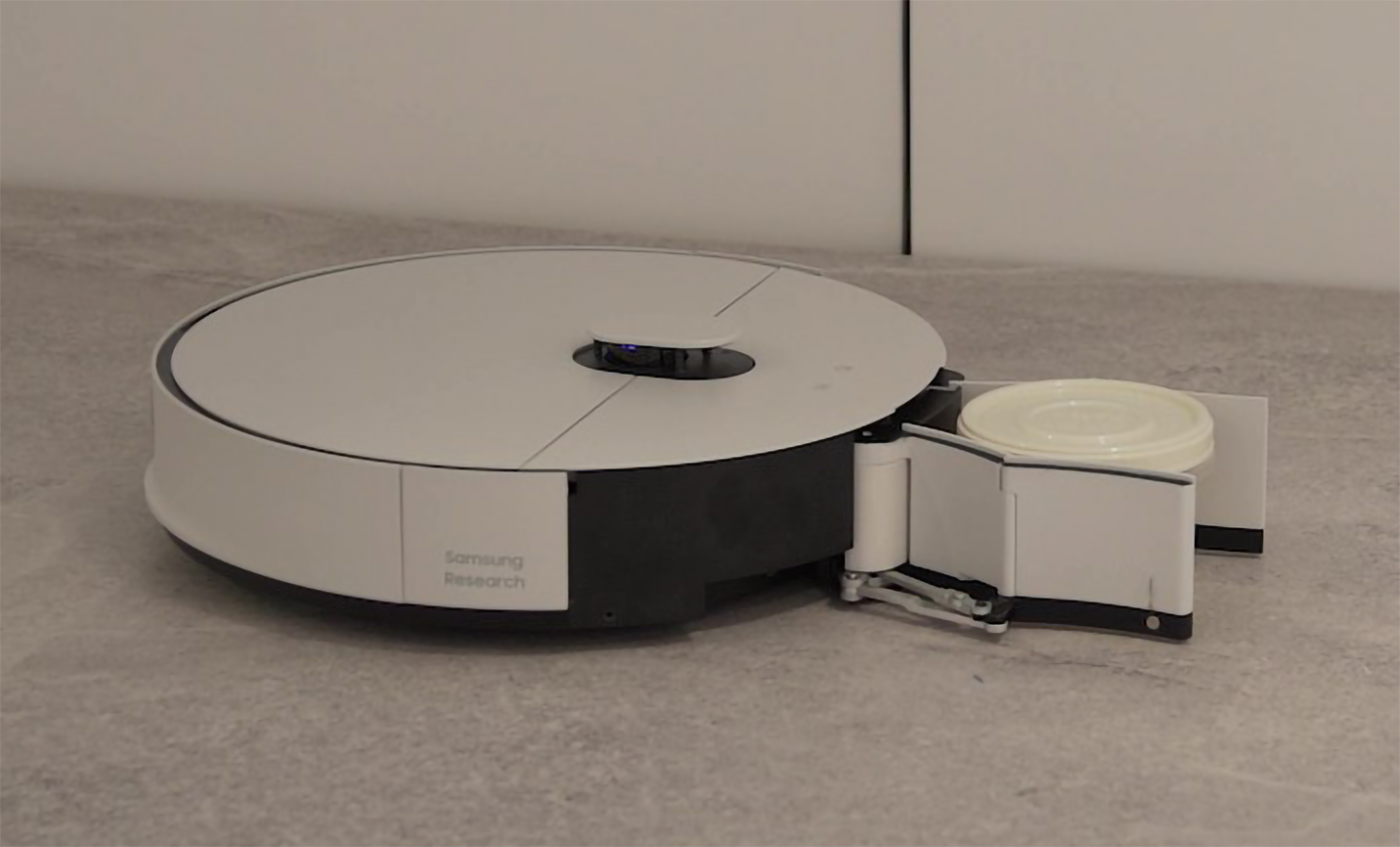}
\caption{A commercial robotic vacuum cleaner equipped with the proposed geometry-induced self-locking deployable linkage. A single deployment actuator transitions the two-stage plates between a backdrivable deployment motion and a near-singular, load-supporting configuration while preserving the original stowed form factor.}
\label{fig1}
\end{figure}

\subsection{Related Work}\label{subsec:related}
Three families of prior work are relevant. \emph{Locking devices} add brakes, clutches, or ratchets to hold a joint once a configuration is reached \cite{plooij2015lock, collins2005bipedal}; Hsu et al.\ \cite{hsu2017self}, for instance, combine a differential with a friction brake so that an underactuated gripper locks automatically once a grasp is formed. Non-backdrivable transmissions such as worm gears or high-friction differentials achieve the same effect intrinsically \cite{controzzi2016sssa}. These are true self-locking solutions, but they add mass, cost, and, for worm gears, efficiency loss in both directions.

\emph{Toggle, over-center, and bistable mechanisms} exploit linkage geometry instead. A toggle clamp obtains a large mechanical advantage as two links approach collinearity and is held by driving the joint slightly past dead-center against a hard stop \cite{norton2020design}; bistable compliant mechanisms store elastic energy to hold either of two states \cite{howell2001compliant, jensen2003identification}. The proposed mechanism belongs to this family but stops \emph{before} dead-center (Section~\ref{subsec:static_analysis}): the plates must fold flush into a curved chassis, which leaves no room for a far-side hard stop or a preload spring.

\emph{Singularity-based stiffness} is a classical principle of closed-loop kinematics: at an input singularity the output loses a degree of freedom and the mechanical advantage becomes unbounded \cite{gosselin1990singularity, zlatanov1995unifying, angeles2003fundamentals}. Closest to our work, Jeon et al.\ \cite{jeon2019singularity} proposed a six-bar toggle linkage for a finger clamping unit that holds its clamping force at an over-singular configuration, using added linkage compliance. Our design differs in its stopping point (before, rather than past, the singular configuration, with no stored elastic energy), its object (a deployable sweeping plate), its topology (a six-bar whose ternary link is the output plate), its release strategy (passive, body-triggered, and kinematically decoupled from the load direction), and its integration into a commercial chassis without modification of the primary structural frame or the stowed external envelope. We claim no new mechanism-theoretic principle; the contribution is the design instantiation and its system-level validation.

Two simpler alternatives deserve comment. A \emph{fixed scoop} cannot be stowed: it would enlarge the footprint, block the bumper and cliff sensors, prevent docking, and obstruct navigation. A \emph{single-stage two-finger gripper} can be stowed, but its sweep width is limited to the plate length, whereas the two-stage linkage roughly doubles the deployed width from the same stowed length; it must hold objects against lateral loads with continuous motor torque or a brake, whereas the proposed linkage holds with the motor de-energized; and a straight plate cannot conform to the curved front housing when stowed. The proposed linkage is the option that satisfies stowability, sweep width, and zero-torque holding simultaneously within the existing chassis volume.

\subsection{Contributions}
\begin{enumerate}
    \item A compact two-stage deployable linkage, driven by a single deployment actuator, that transitions between backdrivable motion and a load-supporting near-singular configuration, integrated into a commercial robotic vacuum cleaner without modifying its primary structural frame, its stowed external envelope, or its navigation pipeline (a second actuator adjusts only the deployment height).
    \item A kinematic and static analysis of the locking condition, consisting of the loop-closure model, the scalar input--output Jacobian, and an analytical holding condition that identifies what resists the residual back-driving torque and how the load direction and manufacturing tolerances enter.
    \item Experimental validation of deployment repeatability (100 cycles), zero-torque load holding (50\,N, in-plane), and pushing-type tidying on seven household objects, with a documented comparison protocol against a commercial arm-equipped vacuum robot.
\end{enumerate}

\section{DESIGN AND OPERATING PRINCIPLE}\label{sec:system_design}

\begin{figure}[!t]
\centering
\begin{subfigure}{0.98\linewidth}
    \centering
    \includegraphics[width=\linewidth]{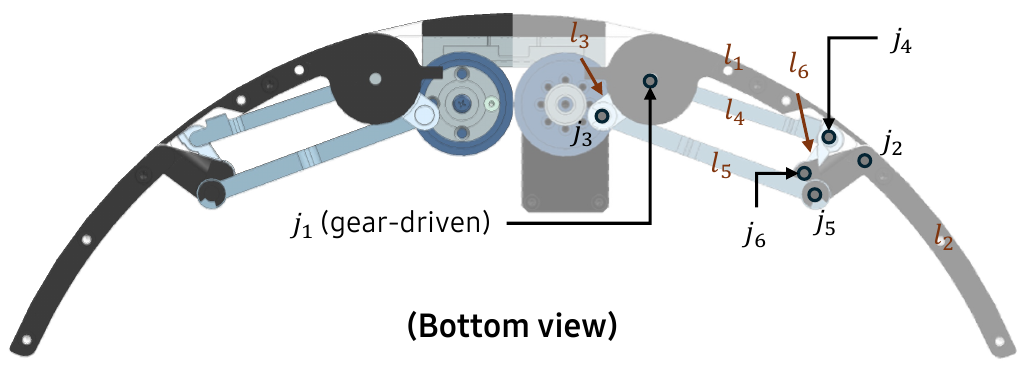}
    \caption{}
    \label{fig2a}
\end{subfigure}
\vfill
\begin{subfigure}[b]{0.98\linewidth}
    \centering
    \includegraphics[width=\linewidth]{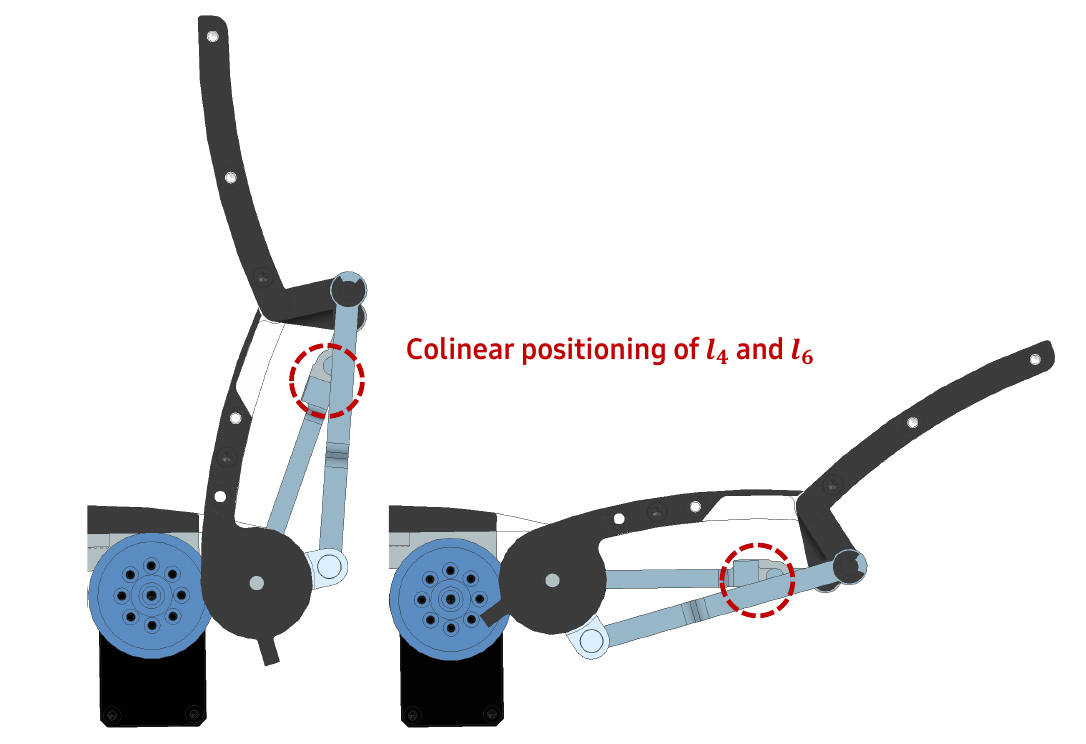}
    \caption{}
    \label{fig2b}
\end{subfigure}
\vfill
\vspace{1.2em}
\begin{subfigure}[b]{0.98\linewidth}
    \centering
    \includegraphics[width=\linewidth]{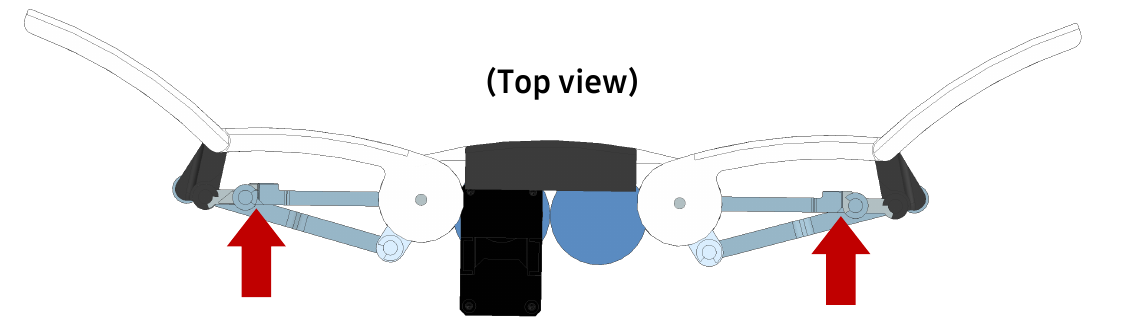}
    \caption{}
    \label{fig2c}
\end{subfigure}
\caption{Operating states of the six-link deployable mechanism. $\theta_{\mathrm{in}}$ is the rotation of the driven joint $j_1$ from the stowed configuration and $\theta_{\mathrm{out}}$ the rotation of $l_2$ relative to the chassis.
(a)~\textit{Backdrivable deployment}: far from alignment the transmission ratio $J$ is $O(1)$; the plates first rotate together about $j_1$ with the internal loop folded, then $l_2$ opens relative to $l_1$.
(b)~\textit{Near-singular rigidification}: the internal subchain $l_4$--$l_6$--$l_2$ approaches collinearity and $J\to 0$, so that plate loads are carried axially along the aligned members.
(c)~\textit{Body-triggered release}: a chassis protrusion perturbs the aligned joint during retraction, restoring $J$ and allowing folding.}
\label{fig:mech_states}
\end{figure}

\subsection{Mechanism Architecture}\label{subsec:architecture}

The mechanism is a six-link, closed-chain planar linkage with links $l_1$--$l_6$ and revolute joints $j_1$--$j_6$ (Fig.~\ref{fig:mech_states}a). The chassis is the fixed frame. Link $l_1$ is the first-stage plate and pivots on the frame joint $j_1$. The deployment actuator does not drive $l_1$ directly: through the gear-train of Section~\ref{sec:implement} it rotates the coaxial pivot at $j_1$ of the internal subchain, which carries $l_1$ along while the internal loop is folded and continues to act on the loop after $l_1$ has stopped. Link $l_2$ is the second-stage plate and a \emph{ternary} link: it connects to $l_1$, to the internal transmission member $l_6$, and to the external structural member $l_3$; the joints to $l_1$ and $l_3$ are coincident, forming the multi-branch joint at the $l_1$--$l_2$--$l_3$ interface. Links $l_4$--$l_6$ couple the frame to $l_2$ through the fixed pivot of $l_5$.

Counting the frame, the chain has seven bodies and eight revolute joints (the coaxial pivots at $j_1$ counted separately), so the Gr\"ubler--Kutzbach count gives $M = 3(7-1) - 2\cdot 8 = 2$. The two degrees of freedom are not used simultaneously. Over the first part of the stroke the internal loop remains folded. No element locks it; because rotating the unloaded plates about $j_1$ requires less torque than opening the internal loop, the driven pivot carries $l_1$ along and the linkage moves as a unit. The travel of $l_1$ is bounded by the shapes of the links and the chassis rather than by a dedicated stop; once $l_1$ reaches this bound, further rotation of the driven joint $j_1$ opens $l_2$ relative to $l_1$ through the internal loop until the subchain $l_4$--$l_2$--$l_6$ approaches collinearity (Section~\ref{subsec:operating_modes}), where a small magnet retains it (Section~\ref{subsec:locking_aids}). In each phase the motion is therefore described by a single coordinate, and what varies along the stroke is the \emph{transmission ratio} from input to second-stage plate.

We use \emph{self-locking} in the sense common to toggle and singularity-based mechanisms \cite{norton2020design, jeon2019singularity}: the linkage geometry, rather than an added brake, resists back-driving in a designated configuration. This is distinct from the intrinsic non-backdrivability of a worm gear \cite{plooij2015lock}. The proposed configuration is \emph{near}-singular and metastable; Section~\ref{subsec:static_analysis} states exactly what holds it, and a small bias element is required against vibration.

\subsection{Loop-Closure Kinematics and Transmission Ratio}\label{subsec:kinematics}

Let $\theta_{\mathrm{in}}$ be the rotation of the driven joint $j_1$ relative to the chassis and $\theta_{\mathrm{out}}$ the rotation of $l_2$ relative to the chassis (Fig.~\ref{fig:mech_states}); both are zero when stowed. In each phase of Section~\ref{subsec:architecture} a single scalar constraint relates the two angles: while the internal loop is folded, $\theta_{\mathrm{out}} = \theta_{\mathrm{in}}$, and once $l_1$ is at the end of its travel, the loop through $l_1$, $l_2$, $l_6$, $l_4$, $l_5$, and the frame yields
\begin{equation}
    f(\theta_{\mathrm{in}}, \theta_{\mathrm{out}}) = 0
    \label{eq:loop}
\end{equation}
and differentiation gives the scalar input--output Jacobian (transmission ratio) along the trajectory,
\begin{equation}
    J(\theta_{\mathrm{in}}) \;\triangleq\; \frac{\mathrm{d}\theta_{\mathrm{out}}}{\mathrm{d}\theta_{\mathrm{in}}}
    = -\,\frac{\partial f/\partial \theta_{\mathrm{in}}}{\partial f/\partial \theta_{\mathrm{out}}}
    \label{eq:jacobian}
\end{equation}
Since $J$ is scalar we refer to $J\to 0$, not to a vanishing determinant. The denominator of \eqref{eq:jacobian} does not vanish in the working range, so $J\to 0$ where the numerator vanishes, i.e., where an incremental input rotation produces no first-order change in loop-closure error. Geometrically this is the configuration in which $l_4$, $l_6$, and the segment of $l_2$ between $j_2$ and $j_4$ are collinear, which is the input-side (Type~I \cite{gosselin1990singularity}) singularity of the constrained motion. Denoting the corresponding input angle by $\theta_{\mathrm{in}}^{\ast}$, a first-order expansion gives
\begin{equation}
    J(\theta_{\mathrm{in}}) \approx k\,(\theta_{\mathrm{in}} - \theta_{\mathrm{in}}^{\ast}), \qquad k = \left.\frac{\mathrm{d}J}{\mathrm{d}\theta_{\mathrm{in}}}\right|_{\theta^\ast}
    \label{eq:linear}
\end{equation}
so that $|J|$ decreases linearly to zero over the last few degrees of travel; on the approach side $\theta_{\mathrm{in}} < \theta_{\mathrm{in}}^{\ast}$, so $J$ and $k$ have the same sign there and only the magnitude $|J|$ matters for what follows.

\subsection{Static Holding Condition}\label{subsec:static_analysis}

The actuator is stopped at $\theta_{\mathrm{in}}^{\max}$, a design margin $\delta\theta = \theta_{\mathrm{in}}^{\ast} - \theta_{\mathrm{in}}^{\max} > 0$ \emph{before} exact collinearity. The mechanism is therefore not an over-center toggle: there is no hard stop past dead-center and the configuration remains reversible. The price is that $J(\theta_{\mathrm{in}}^{\max})$ is small but non-zero. By virtual work, an external moment $\tau_{\mathrm{out}} = F\,r_{\mathrm{tip}}\sin\phi$ on the second-stage plate (tip force $F$ at angle $\phi$ to the plate, lever arm $r_{\mathrm{tip}}$ about $j_2$) is reflected to the actuator side as
\begin{equation}
    \tau_{\mathrm{in}} = J(\theta_{\mathrm{in}})\,\tau_{\mathrm{out}} = J(\theta_{\mathrm{in}})\,F\,r_{\mathrm{tip}}\sin\phi
    \label{eq:residual}
\end{equation}
with mechanical advantage $\mathrm{MA} = 1/J$. The locked state is held with the motor de-energized provided that
\begin{equation}
    |J(\theta_{\mathrm{in}}^{\max})|\,F\,r_{\mathrm{tip}}\sin\phi \;\le\; \tau_f + \tau_m
    \label{eq:holding}
\end{equation}
where $\tau_f$ is the static friction torque of the gear-train and motor reflected to $j_1$ and $\tau_m$ the bias torque of the magnetic detent (Section~\ref{subsec:locking_aids}) projected onto the input coordinate. Equation~\eqref{eq:holding} makes explicit what holds the mechanism: the geometry reduces the demand side by the factor $|J(\theta_{\mathrm{in}}^{\max})| \approx |k|\,\delta\theta$, and ordinary drive-train friction supplies the remainder. The magnet's role is different in kind: friction resists motion but does not restore position, so a small restoring bias is needed to hold the linkage at $\theta_{\mathrm{in}}^{\max}$ under vibration when no task load is present. Its bias torque is that of a small neodymium disc acting over a short lever arm and is not sized to carry task loads. Neither $\tau_f$ nor $\tau_m$ was measured (Section~\ref{subsec:limitations}); the load test of Section~\ref{subsec:robustness}, performed with the motor de-energized, shows empirically that $\tau_f+\tau_m$ exceeds $\tau_{\mathrm{in}}$ at $F = 50$\,N.

The sign of $\tau_{\mathrm{out}}$ decides whether a load drives the subchain toward or away from collinearity. The in-plane task load is the reaction of an object being pushed: it acts on the plate opposite to the sweeping direction, toward the body of the robot. By the sign of \eqref{eq:residual} and the monotonicity of $J$ in \eqref{eq:linear}, this load drives the mechanism \emph{toward} $\theta_{\mathrm{in}}^{\ast}$, where $J$ decreases further and the load is carried axially along the aligned members: the self-reinforcing direction. The load test of Section~\ref{subsec:robustness} applies its force in this same direction (Fig.~\ref{fig5}), so that it characterizes the locked mechanism under the loading it actually experiences in service. An in-plane load of the opposite sign, pulling the plate tip forward and away from the body, would move the input \emph{away} from $\theta_{\mathrm{in}}^{\ast}$ into the region where $J$ grows; it is the direction in which \eqref{eq:holding} is most demanding, but it does not arise in the pushing task considered here and was not tested (Section~\ref{subsec:limitations}). The task and test loads also press $l_1$ against the end of its travel, so they do not back-drive the unit rotation about $j_1$. Out-of-plane loads are not resisted by the singularity but by the bending stiffness of the plates and pins (Section~\ref{sec:discussion}).

Manufacturing tolerances shift $\theta_{\mathrm{in}}^{\ast}$, and by \eqref{eq:linear} a shift $\Delta\theta^\ast$ changes $|J(\theta_{\mathrm{in}}^{\max})|$ by $|k|\,\Delta\theta^\ast$, i.e., by the fraction $\Delta\theta^\ast/\delta\theta$ of its nominal value. The margin $\delta\theta$ was chosen so that typical injection-molding tolerances (of order $0.1$\,mm on link lengths) remain a small fraction of it. Consistently, the measured cycle-to-cycle standard deviation of the locked opening angle, $0.12^\circ$ (Section~\ref{subsec:repeatability}), is an order of magnitude smaller than $\delta\theta$, indicating that the mechanism operates in a repeatable region of the transmission curve rather than on a knife-edge.

\subsection{Operating Modes}\label{subsec:operating_modes}

The stroke is traversed in three regimes (Fig.~\ref{fig:mech_states}). Far from alignment, $J = O(1)$: while the internal loop is folded both plates rotate together about $j_1$ ($J = 1$), and after $l_1$ reaches the end of its travel $l_2$ opens relative to $l_1$ with $J$ still of order one. The linkage is easily backdrivable: an external moment on the plate is reflected to the actuator essentially undiminished by \eqref{eq:residual}. ``Compliant'' is used here only in the sense of low transmission stiffness along the output direction; no claim is made that the robot accommodates external disturbances during deployment, which was not tested.

In the final degrees of travel $J$ falls toward zero, the mechanical advantage rises correspondingly, and loads on the second-stage plate are transmitted axially along $l_4$--$l_6$--$l_2$ into the chassis instead of through the actuator. The actuator is de-energized in this state and the configuration is held by \eqref{eq:holding}.

To leave the rigidified regime, a small protrusion on the chassis engages a release surface on the linkage only when the second-stage plate moves toward the body during commanded retraction. Because this motion is kinematically orthogonal to the in-plane load direction of \eqref{eq:residual}, task loads cannot actuate the trigger. Once perturbed, $J$ grows away from zero and normal coupled folding resumes.

\section{IMPLEMENTATION ON ROBOTIC VACUUM PLATFORM}\label{sec:implement}

The mechanism was integrated into a commercial robotic vacuum platform under the constraints of its existing geometric envelope, low-profile exterior, power budget, and navigation pipeline.

\subsection{Mechanical Integration and Actuation Layout}

\begin{figure}[t]
    \centering
    \includegraphics[width=0.99\linewidth]{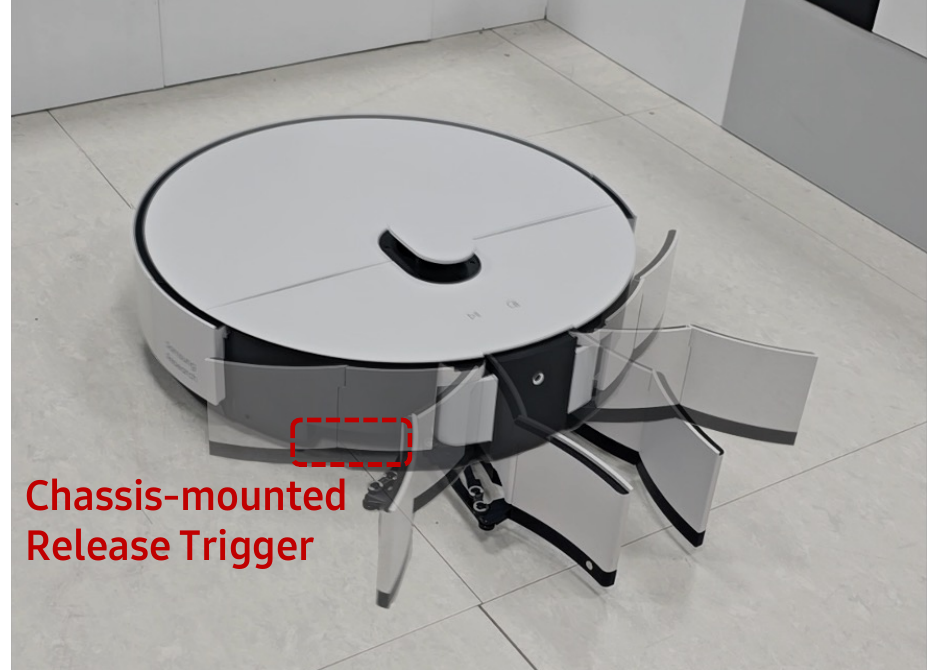}
    \caption{Robotic vacuum with the deployable rigidifying mechanism. Transparent silhouettes show the unfolding and rigidifying sequence as the two-stage plates engage an object; the chassis-mounted release trigger is indicated.}
    \label{fig3}
\end{figure}

Both left and right linkages are driven at $j_1$ by one \emph{deployment actuator} through a symmetric gear-train embedded in the front chassis, ensuring equal rotation of both sides without a second motor or synchronization mechanism. The gear-train fits within the original curved front housing so that the folded plates remain flush with the exterior, and the internal subchain is mounted directly behind the first-stage plates without modifying the primary structural frame (Fig.~\ref{fig3}).

A second, vertically oriented \emph{height-adjust actuator} raises or lowers the whole mechanism relative to the floor to accommodate objects of different thickness. It takes no part in deployment, rigidification, or release; we therefore describe the design as a \emph{single-deployment-actuator} mechanism rather than a single-actuator robot.

\subsection{Deployment Sequence}\label{subsec:sequence}

\begin{itemize}
    \item \textbf{Stowed:} both plates conform to the front curvature; the subchain is folded within the chassis volume.
    \item \textbf{Primary rotation:} both plates rotate outward together through the gear-train with the internal loop folded ($J = 1$).
    \item \textbf{Secondary opening:} once the first-stage plates reach the end of their travel, the second-stage plates open relative to them ($J = O(1)$).
    \item \textbf{Rigidifying transition:} in the final degrees of travel the subchain approaches collinearity; the actuator is stopped at $\theta_{\mathrm{in}}^{\max}$ and de-energized. No hard stop past dead-center is used.
    \item \textbf{Release and folding:} on commanded retraction the chassis protrusion perturbs the aligned joint, restoring mobility, and the plates fold to the stowed state.
\end{itemize}

\begin{figure*}[t]
    \centering
    \begin{subfigure}[b]{0.41\textwidth}
        \centering
        \includegraphics[width=\textwidth]{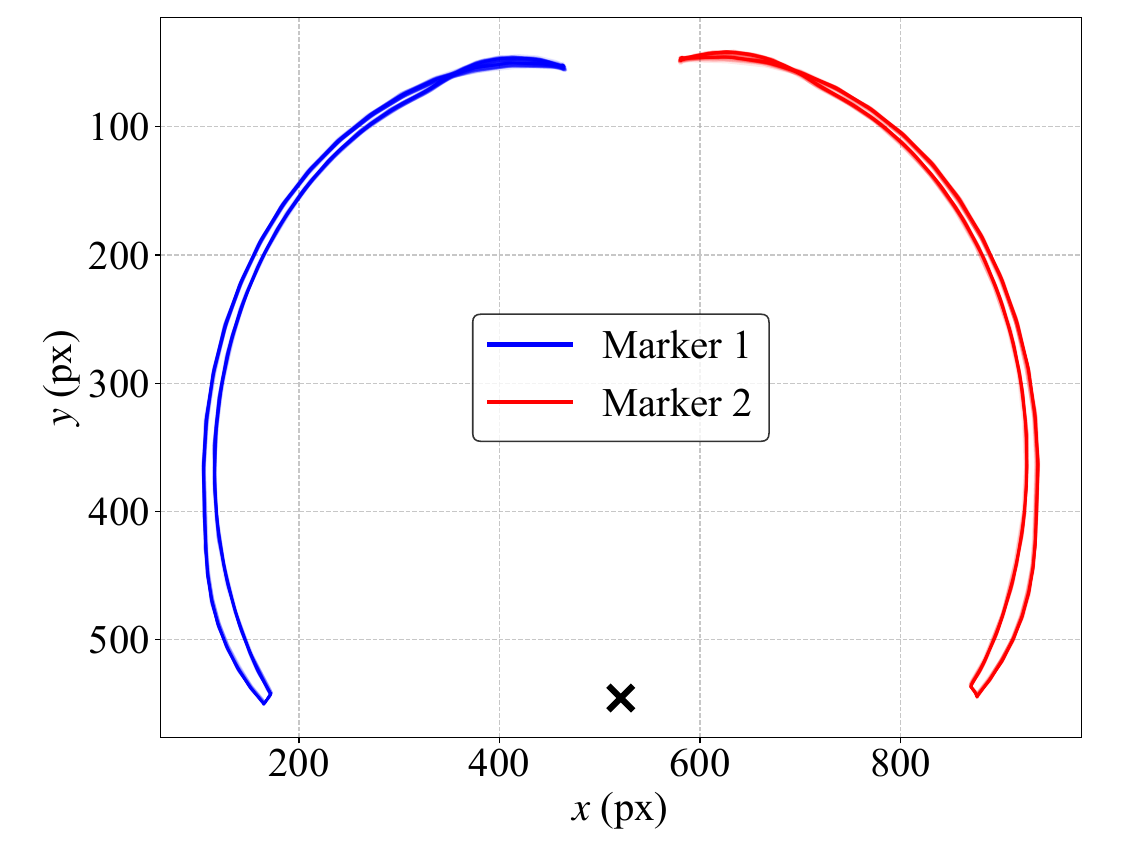}
        \caption{}
        \label{fig4a}
    \end{subfigure}
    \hspace{0.03\textwidth}
    \begin{subfigure}[b]{0.41\textwidth}
        \centering
        \includegraphics[width=\textwidth]{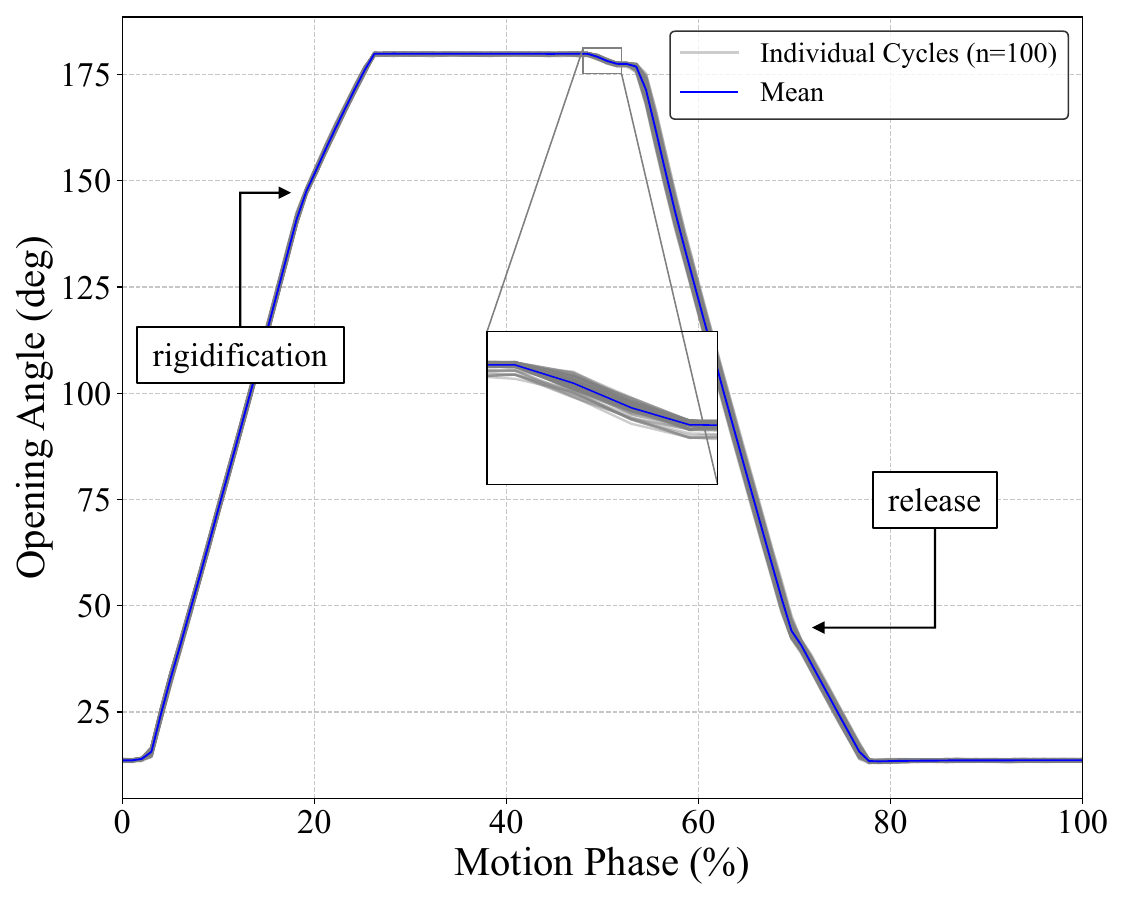}
        \caption{}
        \label{fig4b}
    \end{subfigure}
    \caption{100-cycle repeatability test. (a)~Marker trajectories at the distal ends of the second-stage links in overhead-camera image coordinates; the black cross is the kinematic origin (marker midpoint in the stowed state). (b)~Opening angle between the rays from the origin to the two markers versus normalized motion phase for all 100 cycles; the inset shows the locked state. The opening angle is a monotonic function of $\theta_{\mathrm{in}}$ and $\theta_{\mathrm{out}}$ and is used because it is directly observable from the markers.}
    \label{fig:repeatability}
\end{figure*}

\subsection{Locking Aid and Passive Release}\label{subsec:locking_aids}

A small neodymium magnet is located near the alignment region. Its role, per \eqref{eq:holding}, is to supply the restoring bias $\tau_m$ that keeps the linkage at $\theta_{\mathrm{in}}^{\max}$ under vibration and incidental contact when the actuator is de-energized; it is not the primary load path. Release is achieved passively by the chassis protrusion of Section~\ref{subsec:operating_modes}; no release actuator or sensing is required. Across the 100 repeatability cycles of Section~\ref{subsec:repeatability}, the trigger released the mechanism on every cycle (100/100).

\subsection{Software and Perception Integration}

Integration required no change to the SLAM or navigation pipeline, since the folded configuration preserves the sensing geometry and clearance envelope. A compact forward-facing camera classifies objects coarsely (rigid, deformable, low-profile) and selects a deployment angle for each side before unfolding. This affects only the commanded $\theta_{\mathrm{in}}$, not the locking behavior, which depends on the final configuration alone. Its contribution to the tidying results was not isolated by a fixed-angle ablation (Section~\ref{subsec:limitations}).

\section{EXPERIMENTAL VALIDATION}\label{sec:experiments}

Three experiments were conducted: (1) repeatability of deployment, locking, and release over 100 cycles; (2) load holding of the locked configuration with the motor de-energized; and (3) pushing-type tidying of seven household objects with a documented comparison against two reference systems. Together they address the question: \emph{can a near-singular linkage configuration replace active holding torque for pushing-type tidying on a compact platform?}

\subsection{Repeatability and Release Reliability}
\label{subsec:repeatability}

The robot executed the full sequence of deployment, locking, release, and folding 100 consecutive times. A fixed overhead camera recorded high-contrast markers at the distal ends of the second-stage links; marker centroids were extracted with sub-pixel accuracy and the opening angle was computed about the kinematic origin defined in Fig.~\ref{fig:repeatability}.

Fig.~\ref{fig4a} shows the superposed marker trajectories and Fig.~\ref{fig4b} the phase-folded opening angle. At the locked state the standard deviation of the opening angle across cycles was $0.12^{\circ}$, with no monotonic drift over the 100 cycles. The larger spread during the transient phase (roughly 20--80\,\% of the motion) is attributed to motion blur in the vision tracking rather than to mechanical variability, since the spread collapses in both static states, where the markers are stationary and the tracking noise is at its floor. The release trigger operated on 100 of 100 cycles.

These results show that the linkage converges to the same locked configuration to within a variation an order of magnitude smaller than the design margin $\delta\theta$, consistent with Section~\ref{subsec:static_analysis}. A 100-cycle test is a functional verification, not a durability qualification (Section~\ref{subsec:limitations}).

\subsection{Zero-Torque Load Holding}
\label{subsec:robustness}

\begin{figure}[t]
    \centering
    \includegraphics[width=\linewidth]{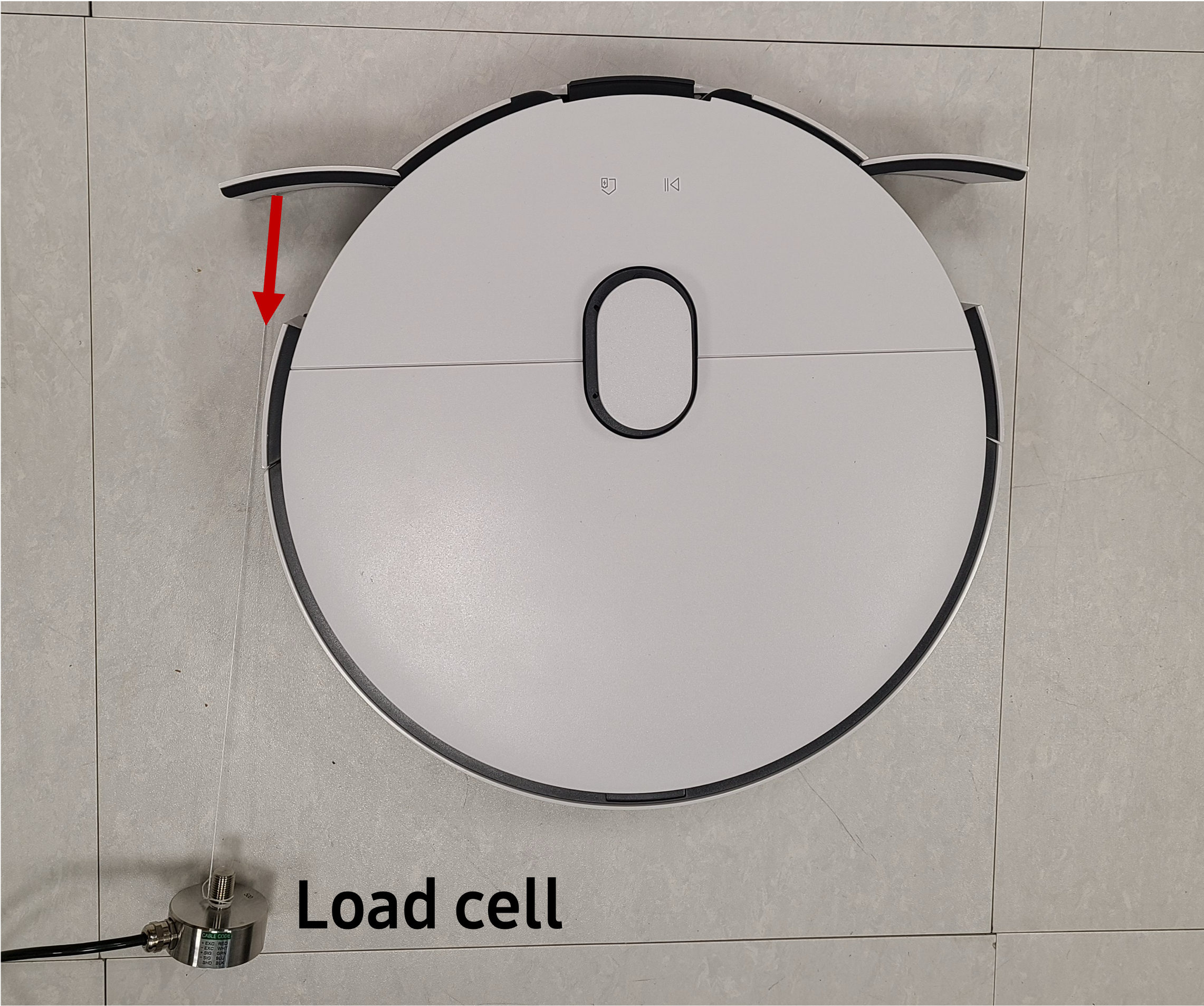}
    \caption{Quasi-static in-plane loading of the locked plate. The chassis is clamped to a fixture; a non-elastic cable runs from the plate tip through an inline tensile load cell, in the direction opposite to sweeping (arrow).}
    \label{fig5}
\end{figure}

The chassis was clamped to a rigid fixture so that it could not translate or rotate (Fig.~\ref{fig5}). The mechanism was deployed to $\theta_{\mathrm{in}}^{\max}$ and the deployment actuator was de-energized, so that only the passive terms of \eqref{eq:holding} could resist back-driving. A non-elastic cable was attached to the tip of the second-stage plate and routed in the mechanism plane in the direction opposite to sweeping, i.e., in the direction of the reaction force from an object being pushed (Section~\ref{subsec:static_analysis}), so that the test reproduces the in-plane loading of the locked plate in service. An inline single-axis tensile load cell in series with the cable measured the applied tension; because the cell is in series, it reads the cable tension along the cable line irrespective of any lateral freedom of the cell body. Tension was increased quasi-statically to 50\,N while the linkage joints were observed.

No unlocking, collapse of the aligned subchain, or relative joint back-driving was observed up to 50\,N, and the actuator shaft remained stationary. This is a pass/fail result at a prescribed load (Section~\ref{subsec:limitations}). The result shows empirically that the passive terms of \eqref{eq:holding} exceed the residual torque of \eqref{eq:residual} at $F=50$\,N.

To put 50\,N in context, pushing a 3\,kg object on a hard floor with a friction coefficient of 0.5 requires about 15\,N of horizontal force; all objects in Section~\ref{subsec:tidying} are household items weighing well under 3\,kg, so 50\,N exceeds the task loads by a substantial margin and is applied in the same direction as the task load; the result therefore transfers directly to the service condition. In-plane loads of the opposite sign, which move the input away from $\theta_{\mathrm{in}}^{\ast}$, do not arise in the pushing task considered here and were not tested (Section~\ref{subsec:limitations}).

Because the actuator was de-energized throughout, the test also demonstrates the state-retention property directly: holding 50\,N required zero electrical power. A backdrivable linkage ($J = O(1)$) would have to supply $\tau_{\mathrm{out}}$ through its drive train for the duration of every pushing maneuver; the near-singular configuration reduces this demand by $1/|J(\theta_{\mathrm{in}}^{\max})|$ and, in practice, to zero. This comparison is analytical (Section~\ref{subsec:limitations}).

\subsection{Tidying Performance and Comparison}
\label{subsec:tidying}

\begin{figure*}[!t]
    \centering
    \includegraphics[width=0.85\linewidth]{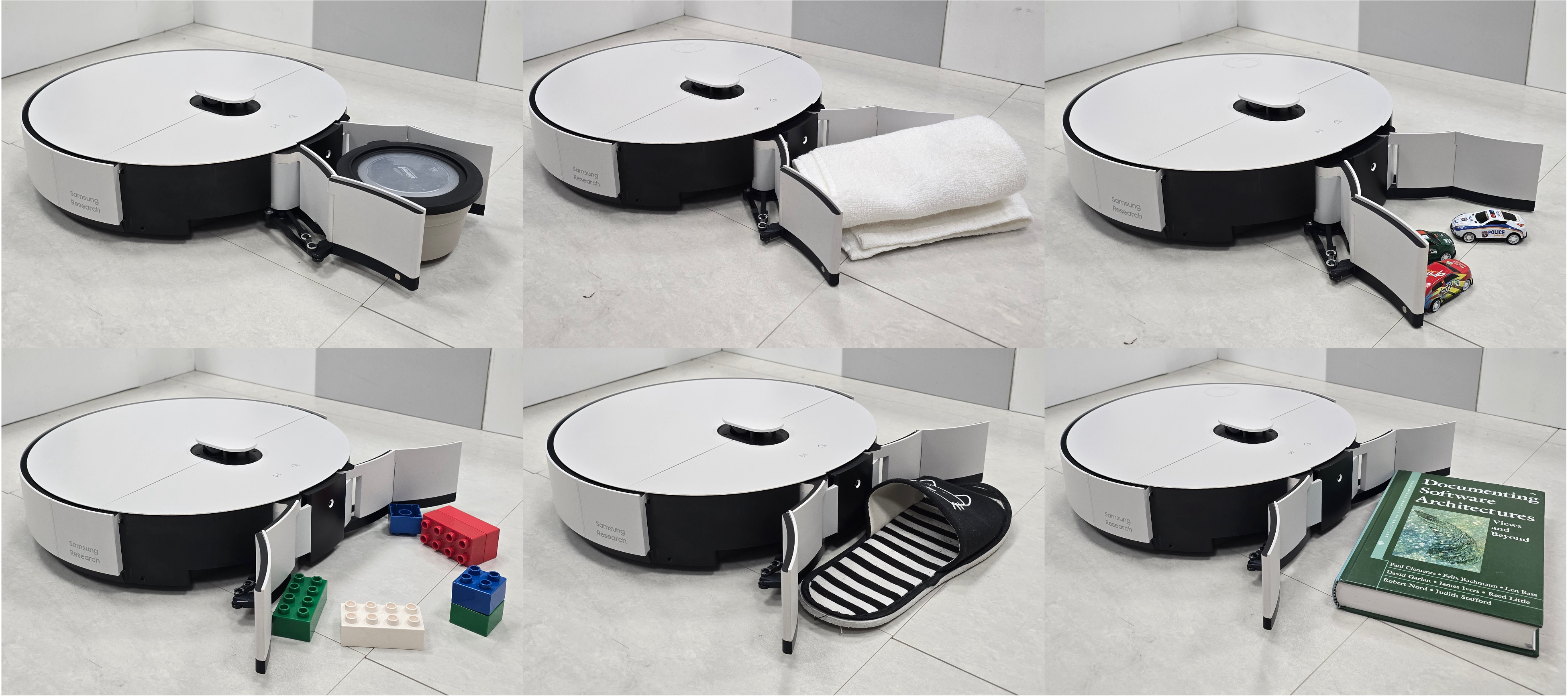}
    \caption{Representative tidying trials on six of the seven test objects (the wadded tissue is omitted for space). Each trial consists of approach, deployment and locking, a single forward push to the target zone, retraction, and departure.}
    \label{fig6}
\end{figure*}

Seven household objects spanning rigid, deformable, bulky, and low-profile categories were used (Table~\ref{tab:tidying_results}). Trials were run on a hard floor in an open area away from walls. For each trial the object was placed at a marked start position; the robot autonomously approached, classified the object, deployed and locked the plates, pushed the object forward over 0.5\,m to a marked target zone, retracted, and reversed. A trial was a \emph{success} if the object's footprint lay entirely within the target zone after the robot withdrew and the object had not escaped the plates or been flipped during transport. Twenty trials were run per object (140 total).

Two commercial systems were run under the same object set, floor, start position, target zone, and success criterion. (1)~An unmodified bumper-only vacuum robot (Samsung VR9700), which has no manipulation capability; it is included not as a competitor but to establish that the base platform cannot perform the task, and its result (0/20 on every object) carries no further information. (2)~A commercial vacuum robot with an integrated multi-DoF arm (Roborock Saros Z70) in its app-commanded pickup mode; each object was presented 20 times, and a trial was a success if the robot grasped and deposited the object inside the target zone. The Z70's task philosophy (grasp and carry) differs from ours (push and contain), so the comparison is confined to the pushing-type relocation task defined above and should not be read as a general capability comparison.

The proposed mechanism succeeded in 140 of 140 trials (Table~\ref{tab:tidying_results}; 95\,\% Clopper--Pearson lower bound 97.4\,\%), each with a single forward motion; representative trials are shown in Fig.~\ref{fig6}. No object escaped laterally, flipped, or jammed; the two-stage plates formed a closed containment pocket around deformable items (towel, tissue) and a stable two-point contact on rigid items (container, textbook). The Z70 succeeded on the wadded tissue and partially on the slipper and toy car, failing on wide, flat, or heavy items; observed failure modes were misaligned grasp, slipping from the gripper during lift, and repeated re-grasp attempts, and successful trials required several grasp, lift, and carry steps.

\begin{table}[!b]
\centering
\caption{Tidying success rate over 20 trials per object for each system. ``Bumper'' is the unmodified base platform (Samsung VR9700); ``Arm'' is the Roborock Saros Z70.}
\label{tab:tidying_results}
\vspace{1mm}
\footnotesize\setlength{\tabcolsep}{3.5pt}
\begin{tabular}{@{}>{\raggedright\arraybackslash}p{2.35cm}|>{\raggedright\arraybackslash}p{2.1cm}|c|c|c@{}}
\hline
Object & Category & Proposed & Bumper & Arm \\
\hline
Plastic toy (Lego blocks) & rigid, small      & 100\% & 0\% & 0\% \\
Container (Lock\&Lock)     & rigid, bulky      & 100\% & 0\% & 0\% \\
Toy car (die-cast)         & rigid, small      & 100\% & 0\% & 20\% \\
Towel                      & deformable, wide  & 100\% & 0\% & 0\% \\
Tissue (wadded)            & deformable, light & 100\% & 0\% & 100\% \\
Indoor slipper             & low-profile       & 100\% & 0\% & 50\% \\
Textbook (hard cover)      & rigid, flat, heavy& 100\% & 0\% & 0\% \\
\hline
\end{tabular}
\end{table}

The 100\,\% figure should be read within its protocol: an open hard floor, a fixed transport distance, objects light enough that the required push force is far below the 50\,N holding capacity, and objects chosen to span categories rather than sampled from real homes. Conditions not tested, under which failures are plausible, include pushing along walls and into corners, carpet, objects taller than the plate or narrower than the gap between the stowed plates, and releasing the object at the target without dragging it back on retraction.

\section{DISCUSSION}\label{sec:discussion}

\subsection{Summary}

The linkage reaches the same near-singular configuration on every cycle, holds an in-plane load of 50\,N in the task direction with the motor de-energized, and completes a well-defined pushing-type tidying task on objects that a bumper-only platform cannot move at all and that a compact arm-equipped platform handles only partially. Equation~\eqref{eq:holding} explains the load result: the geometry reduces the actuator-side demand by $1/|J(\theta_{\mathrm{in}}^{\max})|$ and ordinary drive-train friction absorbs the remainder. The mechanism is not self-locking in the way a worm gear is; it is metastable near a singularity, and we retain the term only in the sense standard for toggle and singularity-based linkages.

The principle that a closed chain near an input singularity resists back-driving is classical; what is new is the design- and system-level instantiation summarized in Section~\ref{subsec:related}, shown to fit within the stowed envelope of a commercial chassis without modifying its primary structural frame. We regard the mechanism as an intermediate design point between passive compliance and articulated manipulation, not as a replacement for grasping.

A rigid plate deployed at floor level introduces pinch points at the hinges and an impact surface at the plate edges. The plate is rigid only when fully deployed and the deployment motion is backdrivable, the deployment force is bounded by the actuator current limit, and the bumper and cliff sensors remain active with the plates deployed. The deployed plates nevertheless enlarge the footprint and should be treated as a moving part under ISO~13482 \cite{iso13482}; people-detection-gated deployment and a compliant plate edge would be required before consumer deployment.

\subsection{Limitations and Future Work}\label{subsec:limitations}

Several aspects were not addressed experimentally and bound the claims of this paper. The contribution of the magnet has been argued analytically but not isolated by an ablation, and the load test established only that the locked plate holds 50\,N, not the force--displacement behavior or the load at which unlocking occurs. Only in-plane, quasi-static loading in the task direction with a fixed chassis was tested; loads of the opposite in-plane sign, off-axis and out-of-plane loads, and dynamic contacts remain to be characterized, and off-axis loads are in any case resisted by plate and pin bending rather than by the singularity, which bounds the vertical loads the plates can carry when partially lifting an object. Motor current in a torque-holding configuration was not measured, so the energy comparison is analytical. The tidying evaluation used a curated object set on an open hard floor without a fixed-angle ablation of the perception-based angle selection; walls, corners, carpet, and release-at-target behavior were not tested. Finally, 100 cycles is a functional verification, not a durability qualification: wear-related drift of $\theta_{\mathrm{in}}^{\ast}$, which directly erodes the margin in \eqref{eq:holding}, and sensing of the locked state are the first items for future work, together with an ablation with the magnet removed and measurement of the unlocking load and tip stiffness, followed by extension of the geometric locking principle to multi-directional stiffness.

\section{CONCLUSION}\label{sec:conclusion}

This paper presented a two-stage deployable linkage for a robotic vacuum cleaner that obtains a load-bearing configuration from its own geometry. By driving an internal three-link subchain to a small margin before collinearity, the transmission ratio from actuator to plate approaches zero, so that task loads are carried through the aligned members into the chassis and the residual actuator-side torque is absorbed by drive-train friction, with a lightweight magnet holding the configuration against vibration. A single deployment actuator performs deployment and locking, and a passive body-triggered release restores mobility for retraction. The design was validated by a 100-cycle repeatability test, a 50\,N zero-torque hold, and 140 pushing-type tidying trials under a documented comparison protocol. Several experiments needed to characterize its limits remain to be done; within its intended scope, it offers a compact way, with a single deployment actuator and few added components, to give a commercial cleaning robot a useful, rigid interaction mode.

\bibliographystyle{IEEEtran}
\bibliography{ref}

\end{document}